\documentclass[conference]{IEEEtran}
\IEEEoverridecommandlockouts
\usepackage{fancyhdr}
\usepackage{amsthm}
\usepackage{graphics} % for pdf, bitmapped graphics files
\usepackage[pdftex]{graphicx}
\usepackage{epsfig} % for postscript graphics files
\usepackage{amsmath} % assumes amsmath package installed
\usepackage{amssymb}  % assumes amsmath package installed
\usepackage{floatrow}
\newfloatcommand{capbtabbox}{table}[][\FBwidth]

\usepackage{blindtext}
\usepackage{mathrsfs}
\usepackage{subfigure}
\usepackage{cite}
\usepackage{color}
\usepackage{nomencl}
\usepackage[automake]{glossaries}
\usepackage{makeidx}
\makenomenclature
\makeglossaries

{}
\usepackage{url}
\usepackage{indentfirst}
\usepackage{palatino}
\usepackage{tikz}
\usetikzlibrary{shapes.geometric, arrows}
\graphicspath{{figures/}}
\usepackage{etoolbox}
\usepackage{algorithm}  
\usepackage{algpseudocode}  
\usepackage{rotating} 
\usepackage{xcolor}
\usepackage{float}
\usepackage{caption}
\usepackage{multirow}
\usepackage{longtable}
\usepackage{epstopdf}
\usepackage{algorithmicx}
\usepackage{booktabs}
\usepackage{amsfonts}
\usepackage{bm}
\makeatletter
\def\BState{\State\hskip-\ALG@thistlm}
\makeatother
\glsaddall
\usepackage[absolute,overlay]{textpos}

\newacronym{TRPO}{TRPO}{Trust Region Policy Optimisation}
\newacronym{LSTM}{LSTM}{Long Short-Term Memory}
\newacronym{RNN}{RNN}{Recurrent Neural Network}
\newacronym{ANN}{ANN}{Artificial Neural Network}
\newacronym{DANN}{DANN}{Distributed Artificial Neural Network}
\newacronym{DBN}{DBN}{Deep Belief Network}
\newacronym{SVR}{SVR}{Support Vector Regression}
\newacronym{FL}{FL}{Federated Learning}
\newacronym{PPO}{PPO}{Proximal Policy Optimisation}
\newacronym{DPPO}{DPPO}{Distributed Proximal Policy Optimisation}
\newacronym{MDP}{MDP}{Markov Decision Process}
\newacronym{CHP}{CHP}{Combined Heat and Power}
\newacronym{BESS}{BESS}{Battery Energy Storage Systems}
\newacronym{TESS}{TESS}{Thermal Energy Storage Systems}
\newacronym{DSM}{DSM}{Demand Side Management}
\newacronym{PV}{PV}{Photovoltaics}
\newacronym{GWO}{GWO}{Grey Wolf Optimisation}
\newacronym{TVAC-PSO}{TVAC-PSO}{Time Varying Acceleration Coefficients Particle Swarm Optimisation}
\newacronym{ADMM}{ADMM}{Alternating Direction Method of Multipliers}
\newacronym{MAS}{MAS}{Multi-Agent Systems}
\newacronym{SOC}{SOC}{State Of Charge}
\newacronym{DDBN}{DDBN}{Distributed Deep Belief Networks}
\begin{document}
% \title{Short-term Load Forecasting with Distributed Long Short-Term Memory}

% \author{\author{\IEEEauthorblockN{1\textsuperscript{st} Yang Chen}
% \and
% \IEEEauthorblockN{2\textsuperscript{nd} Xingyu Zhao}
% \IEEEauthorblockA{\textit{Department of Computer Science} \\
% \textit{University of Liverpool}\\
% Liverpool, the UK \\
% xingyu.zhao@liverpool.ac.uk}}

\title{Short-term Load Forecasting with Distributed Long Short-Term Memory\\
\thanks{This work is supported by the UK EPSRC through End-to-End Conceptual Guarding of Neural Architectures [EP/T026995/1]).}
}
\author{
\IEEEauthorblockN{Yi Dong}
\IEEEauthorblockA{\textit{Department of Computer Science} \\
\textit{University of Liverpool}\\
Liverpool, the UK \\
yi.dong@liverpool.ac.uk}\vspace{-24pt}
\and
\IEEEauthorblockN{Yang Chen}
\IEEEauthorblockA{\textit{Department of R \& D}\\\textit{Suzhou SeeEx Technology Co., Ltd} \\
Jiangsu, China \\
yang.chen@seeextech.com}\vspace{-24pt}
\and
\IEEEauthorblockN{Xingyu Zhao, Xiaowei Huang}
\IEEEauthorblockA{\textit{Department of Computer Science} \\
\textit{University of Liverpool}\\
Liverpool, the UK \\
\{xingyu.zhao, xiaowei.huang\}@liverpool.ac.uk} \vspace{-24pt}
}

\begin{textblock*}{20cm}(1cm,1cm)
	\textcolor{red}{{\large Accepted by 2023 IEEE ISGT North America. To appear in IEEE Xplore.}}
\end{textblock*}

\maketitle
\begin{abstract}
With the employment of smart meters, massive data on consumer behaviour can be collected by retailers.
From the collected data, the retailers may obtain the household profile information and implement demand response. While retailers prefer to acquire a model as accurate as possible among different customers, there are two major challenges. First, different retailers in the retail market do not share their consumer’s electricity consumption data as these data are regarded as their assets, which has led to the problem of data island. Second,  the electricity load data are highly heterogeneous since different retailers may serve various consumers. 
% From the collected data, the retailers may obtain the household profile information and implement demand response. 
% However, different retailers in the retail market do not share their consumer’s electricity consumption data as these data are regarded as their assets, which has led to the problem of data island. 
% Meanwhile, the electricity load data are highly heterogeneous since different retailers may serve various consumers and retailers prefer to acquire an accurate model among different customers as much as possible. 
To this end, a fully distributed short-term load forecasting framework based on a consensus algorithm and \gls{LSTM} is proposed, which may protect the customer’s privacy and satisfy the accurate load forecasting requirement. Specifically, a fully distributed learning framework is exploited for distributed training, and a consensus technique is applied to meet confidential privacy.
Case studies show that the proposed method has comparable performance with centralised methods regarding the accuracy, but the proposed method shows advantages in training speed and data privacy.

\end{abstract}


% \fi
\IEEEpeerreviewmaketitle

\section{Introduction}
Electricity load forecasting is an essential basis for not only industrial production but also social life. The analysis of these load data could help in revealing household profile information and enabling other uses \cite{fallah2019computational,hippert2001neural,taylor2007short}. Specifically, retailers could obtain the electricity consumption behaviour of consumers and provide social and behavioural incentive signals to optimise customers' electricity usage. However, customers' data are not able to be shared among different power supply companies. Therefore, an appropriate algorithm that can generate accurate short-term load forecasting (STLF) while protecting custom privacy is essential for power system operation.

Load forecasting has been an essential part of the field of power system research since the 1970s, starting with some of the earliest methods, such as linear regression (LR) \cite{papalexopoulos1990regression}, stochastic time series (STS) \cite{hagan1987time} and general exponential smoothing (GES) \cite{christiaanse1971short}. The mean accuracy of these methods is about 90\%$\sim$96\%. 
With the introduction of machine learning and artificial intelligence methods, scholars started applying artificial intelligence algorithms to STLF and achieved great success, which increased the accuracy to more than 97\% \cite{lu1993neural,hippert2001neural,kiartzis1995short}.

Due to the vigorous development of hardware equipment and the superb performance of deep learning in terms of the characteristics of nonlinear functions, deep learning algorithms have become particularly popular in load prediction research in recent years.
\cite{deng2019multi} proposed a novel model multi-scale convolutional neural network with time-cognition (TCMS-CNN), which combines sufficient and discriminative features to extract potential law in the dataset providing an excellent result.
\cite{alhussein2020hybrid} proposed a deep learning framework based on a combination of a convolutional neural network (CNN) and long short-term memory (LSTM), which could provide a significant improvement in the accuracy of individual household load forecasting.
\cite{kong2019improved} improved deep belief networks (DBN) with Gauss-Bernoulli restricted Boltzmann machine (GB-RBM) and gray relational analysis (GRA). The developed method has a better performance than DBN and other traditional methods.
However, the dataset required by the aforementioned methods includes all customers' data together without considering individual customers' privacy.

With the rapid development of big data, customer data privacy has gradually received attention. Over the last few years, several governments also have committed to data privacy protection, e.g., the European Commission's General Data Protection
Regulation \cite{voigt2017eu} and the Consumer Privacy Bill of Rights in
the US \cite{privacypolicy}. As a result, \textit{decentralised} learning methods, such as federal learning, have been proposed to solve these problems.
\cite{fekri2022distributed} proposed a federated learning method for load forecasting with smart meter data that is capable of training a machine learning model in a distributed manner without requiring the participant to share their local data.
\cite{dong2021short} proposed a fully distributed STLF method based on \gls{DDBN}, which can solve the STLF model by local computing agents (CA) and update the model parameters by communicating with connected neighbours.
However, these distributed learning algorithms still require a data centre to communicate with individual nodes to update global parameters. In the case of huge data volumes, the central-based distributed method could cause traffic jams and scalability issues.

To this end, a novel fully distributed LSTM-based STLF method is proposed in this paper. It can be trained with a local dataset and converge to global optima by communicating with its connected neighbours, which does not need to transfer any customers' data. Different from the federated learning framework, the proposed consensus-based framework does not have a single computing centre to deal with an integrated global model among all local agents, and therefore the proposed algorithm will not meet computing burdens. Apart from data privacy and communication congestion, the distributed training framework can also avoid the local over-fitting problem since its parameters will be amended based on its neighbouring information.

The major contributions of this paper include:
\begin{enumerate}
    \item A novel distributed LSTM algorithm is proposed for STLF, which can train the model at local and only communicate with its connected neighbours, preserving data privacy.
    \item Due to the parallel computational framework, the proposed distributed method can significantly reduce the training time.
    \item The over-fitting problem of the local model can be mitigated by mutual correction during the communication and training process.
\end{enumerate}

The rest of the paper is organised as follows. Mathematical preliminaries applied in this paper are summarised in Sec. \ref{sec.preliminaries}. The proposed distributed LSTM method is described in Sec. \ref{sec.DLSTM}. Simulation results with discussion are presented in Sec. \ref{sec.simulation}. Finally, we conclude the work in Sec. \ref{sec_con}.

\section{Applied Methodologies}\label{sec.preliminaries}

In this section, we recall methods regarding graph theory, consensus control and LSTM. Let $\mathbb{R}^{n\times m}$ be the set of $n\times m$ real matrices and the superscript $T$ means the transpose of real matrices. $I_N$ denotes the identity matrix of dimension $N$ and $\mathbf 1_N$ represents a column vector with all entries being 1. $\mathbb{R}^{++}$ denotes the positive real numbers. $\|\cdot\|_2$ represents the 2-norm of the argument.%, \textcolor{blue}{which is two dimensional Euclidean space}.%$1_N$ denotes the $N$-dimension column vector with all entries 1.%$A\otimes B$ denotes the Kronecker product of matrices $A$ and $B$.
\subsection{Graph Theory}

Following \cite{biggs1993algebraic}, an undirected graph $\mathcal{G = (V, E)}$ can be used to describe the communication topology among the local computing centres, where $\mathcal{V}=\{\nu_{1},\cdots,\nu_{N}\}$ is the vertex set and $\mathcal{E}\in \mathcal{V}\times\mathcal{V}$ is the edge set. The adjacency matrix $\mathcal{A}=[a_{ij}]\in \mathbb{R}^{N\times N}$ of $\mathcal{G(V,E)}$ is an $N\times N$ matrix, such that $a_{ij} = 1$ if $(\nu_{j},\nu_{i})\in \mathcal{E}$ and $a_{ij}=0$ otherwise. Define the degree matrix $D=\text{diag}\{\sum_{j=1}^{N}a_{1j},\sum_{j=1}^{N}a_{2j},$ $\cdots,\sum_{j=1}^{N}a_{Nj}\}$. A graph is connected if and only if every pair of vertices can be connected by a path, namely, a sequence of edges. In this paper, we assume that the graph is connected and undirected. The Laplacian matrix related to $\mathcal{G(V,E,A)}$ is defined as $\mathcal{L=D-A}$, i.e.,
\small\begin{equation}\label{laplace}
	\mathcal{L}=\left\{\begin{array}
		ll_{ij}=-a_{ij}, i\neq j\\l_{ii}=\sum_{i\neq j}a_{ij}.
	\end{array}\right.
\end{equation}\normalsize
%For a directed graph $\mathcal{G(V,E,A)}$ which is strongly connected, the Laplacian matrix $\mathcal{L}$ is irreducible and satisfies that $\sum_{j=1}^{N}l_{ij}=0$ \cite{lv2017novel}, 0 is an eigenvalue of $\mathcal{L}$ \cite{ren2005consensus} and $1_nr^T=0$ \cite{li2015distributed}, where $r=[r1,r2,\cdots,r_N]$ is the positive left eigenvector of the Laplacian matrix $\mathcal{L}$ associated with the zero eigenvalue.
%\begin{lem}
%Let $\mathcal{G}$ be a graph. Then the dimension of the null space of $\mathcal{L(G)}$ is the number of connected
%components of $\mathcal{G}$.	
%\end{lem}
%Suppose that $x\in \mathbb{R}^N $is an eigenvector of $\mathcal{L(G)}$ corresponding to eigenvalue 0, i.e. that $\mathcal{L(G)}x=0$. Then,
%\begin{equation}
%	x^T\mathcal{L}x=\sum_{i,j\in\mathcal{E}}(x_i-x_j)^2=0.
%\end{equation}

When $\mathcal{G(V,E)}$ is a connected undirected graph, 0 is an eigenvalue of Laplacian $\mathcal{L}$ with the eigenvector $\mathbf 1_N$ and all the other eigenvalues are positive. Then,
\small\begin{equation}\label{nullspace}
	\mathcal{L}1_N=0_N,\ \ 1_N^T\mathcal{L}=0_N^T.
\end{equation}\normalsize

In this paper, we assume that the connection between computing centres are undirected since the communication line has no direction.

\subsection{Long Short-Term Memory}
\begin{figure}
    \centering
    \includegraphics[width=0.8\hsize]{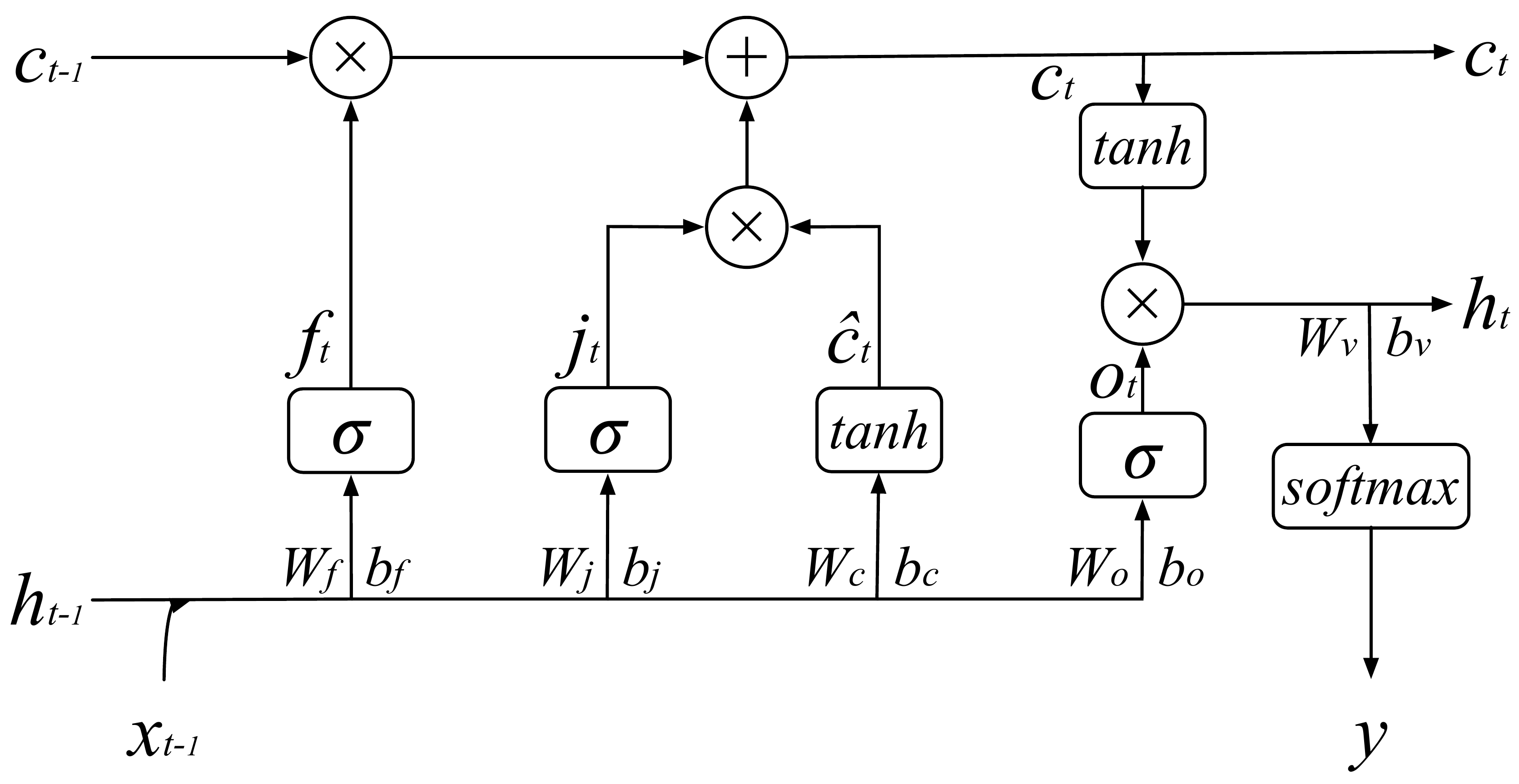}\vspace{-6pt}
    \caption{Structure of LSTM.	\vspace{-12pt}}
    \label{LSTM}
\end{figure}

The LSTM is a refined model of \gls{RNN}, and it has the advantage of avoiding exploding and vanishing gradient problem \cite{van2020review}. The basic structure of an LSTM network is shown in Fig.~\ref{LSTM}.

The propagate steps of LSTM can be separated into four sub-steps: $forget\ gate$ step, $input\ gate$ step, $update$ step and $output$ step.
The $forget\ gate$ step is made by a sigmoid layer which is to decide what information we are going to forget. The inputs of this step are historical output $h_{t-1}$ and current state $x_t$, and it outputs a number within a range $(0,1)$.
\small\begin{equation}
    f_t = \sigma(W_f\cdot[h_{t-1},x_t] + b_f)
\end{equation}\normalsize
where $W_f, b_f$ are the weights and bias of the LSTM forget gate layer, and $\sigma$ denotes the sigmoid activation function.

The second step is composed of a sigmoid layer and a tanh layer. This sigmoid layer is to decide what new information we are going to memorise, and the tanh layer is to estimate the current cell state $\hat{C_{t}}$:
\small\begin{align}
    j_t =& \sigma(W_j\cdot[h_{t-1},x_t] + b_j)\\
    \hat{C_{t}} =& \tanh(W_C\cdot[h_{t-1},x_t] + b_C)
\end{align}\normalsize
The third step is to combine the outputs of second step and update the cell state:
\small\begin{equation}
    C_t = f_t \times C_{t-1} + j_t\times\hat{C_{t}}
\end{equation}\normalsize

Finally, the LSTM yields the output controlled by the $output$ step, which is based on the current cell state $C_t$, history output $h_{t-1}$ and current input $x_t$. It uses a tanh activate function as a filter to push the cell state between $-1$ and $1$, so that the $output$ step can only output the parts we choose.
\small\begin{align}
    o_t =& \sigma(W_o\cdot[h_{t-1},x_t] + b_o)\\
    h_{t} =& o_t \times\tanh(C_t)
\end{align}\normalsize

There is the final output $y$ if end of the hidden layers:
\small\begin{equation}
    y = \textit{softmax}(W_v\cdot h_{t} + b_v)
\end{equation}\normalsize

\section{Distributed Long Short-Term Memory}\label{sec.DLSTM}
In this section, we formulate a Distributed Long Short-Term Memory (DLSTM) method for STLF by consensus-based approaches. It consists of a group of $N$ local LSTM models distributed over a connected graph, where each local computing centre has its own local dataset and cannot be revealed to other computing centres. The objective of the LSTM models is to minimise the empirical loss over the entire data set, which is formulated as
\small\begin{equation}
	\min_{\Theta\in\mathbb{R}^n}\mathbb{E}(\mathcal{D},\Theta) = \sum_{i=1}^{N}\mathbb{E}_i(\mathcal{D}_i,\Theta_i)
\end{equation}\normalsize
where $\mathcal{D}_i$  and $\Theta_i\in\mathbb{R}^n$ are the sub-dataset and weight parameters of the $i$th local LSTM model agent, respectively. $n$ is the dimension of the weight matrix. %In this paper, we assume that the local sub-datasets are sampled from a single optimal model. This is one of the most common assumption in machine learning problems, for example, \textcolor{red}{WAITING REFERENCES...}.
It is noticed that the weight parameters $\Theta_i$ includes all the parameters in LSTM models, include $forget\ gate$, $input\ gate$, $update$ and $output$ steps:
\small\begin{equation}
    \Theta_i = [W_f,b_f,W_j,b_j,W_C,b_C,W_o,b_o]^T
\end{equation}\normalsize

In the traditional LSTM algorithm, the training process is to calculate the gradient and update all the weights $\Theta$:
\small\begin{equation}
    \Theta^t = \Theta^{t-1}-\eta_i\nabla_\Theta\mathbb{E}(\mathcal{D},\Theta^{t-1})
\end{equation}\normalsize

For the DLSTM algorithm, the gradient descent method is applied to optimise the parameters of local LSTM models, and the distributed consensus algorithm is applied to optimise the parameters between connected LSTM models. All the communication and training processes only transfer the weights of different models, which does not require any customer data, and therefore the privacy can be protected. The procedure code of DLSTM is shown in Algorithm \ref{euclid}.

\begin{algorithm}[h]
\small
	\caption{Distributed Long Short Term Memory.}\label{euclid}
	\begin{algorithmic}[1]
		\Procedure{DLSTM}{LSTM, Agents}
		\State $\textit{LSTM} \gets \text{Initial weights $\Theta_i$ for local LSTM models}$
		\State $\textit{Sites} \gets \text{Array of Network pipes to local CA$_i$}$
		\While {$\textit{Sites}\ \text{contains unused data}$}
		\State $\textit{//Start local training for each local site}$
		\For {$\text{each }$ CA$_i$$\in\textit{Sites}$}
		\State {CA$_i$$\gets\text{Forward Propagation}$}
		\State {CA$_i$$\gets\text{Gradient Descent}$} Calculation
		\State {CA$_i$$\gets\text{Back Propagation}$ with \eqref{consensus}}
		\EndFor
		\State $\textit{//Consensus the weight of local LSTM models}$
		%		\State $\text{\boldsymbol{elseif}}$
		\While {$\text{weights are different}$} 
% 		\State {Switching the topologies by transition probability $\gamma$}
% 		\State {Apply the typical Laplacian matrix}
		\For {each CA$_i$ $\in\textit{Sites}$}
		\State {$\text{Consensus weight}$ with \eqref{consensus}}
		\EndFor
		\EndWhile
		\State $\textit{DLSTM}\gets\text{Consensus weight}$
		\EndWhile
		\EndProcedure
	\end{algorithmic}
\end{algorithm}

During each training step of the local computing agent, we add an communication update step for each local LSTM model. Thus, the gradient-based training of DLSTM can be formulated as the following sub-steps:
\small\begin{align}
	\Phi_i^t &= \Theta_i^{t-1}-\eta_i\nabla_\Theta\mathbb{E}_i(\mathcal{D}_i,\Theta_i^{t-1})\label{localtrain}\\
	\Theta_i^t &= \sum_{j\in\mathbb{N}}a_{ij}\Phi_i^t\label{consensus}
\end{align}\normalsize
where $\Phi_i^t$ is the intermediate variable and $\eta_i\in[0,1]$ is the learning rate of the local model. $a_{ij}$ is the element of Laplace matrix $\mathcal{L}$ in \eqref{laplace}. The updates of weights $\Theta_i$ include two procedures, which make up the learning before consensus (LBC) algorithm. In the first stage, the neural networks are trained independently with local sub-dataset through the gradient descent algorithm \eqref{localtrain}.  This stage only uses local information, such as the weights of latest step $\Theta_i^{t-1}$ and the gradient of the empirical loss $\nabla_\Theta\mathbb{E}_i(\mathcal{D}_i,\Theta_i^{t-1})$. In the second stage, the weights of the neural networks are updated via consensus algorithm, where neighbouring information $\Phi_i^t$ are applied. With this learning process, all the agents are able to obtain the single and optimal neural network model when $t$ tends to $\infty$ \cite{li2021cooperative}. %, which will be demonstrated by the theoretical convergence analysis. 
Similarly, the equations \eqref{consensus} and \eqref{localtrain} can be swapped to yield the consensus before learning (CBL) algorithm as 
\small\begin{align}
\Phi_i^t &= \sum_{j\in\mathbb{N}}a_{ij}\Theta_i^{t-1}\label{consensus1}\\
\Theta_i^t &= \Phi_i^{t}-\eta_i\nabla_\Theta\mathbb{E}_i(\mathcal{D}_i,\Phi_i^{t})\label{localtrain1}
\end{align}
\normalsize
Note that the CBL algorithm is to consensus the gradients first, then update the local model. In contrast, the LBC algorithm is to update the local model first, then consensus the models. Theoretically, the gradients will converge to $0$, and the model will become the optimal model with the training time goes infinity. Therefore, there is not much difference between CBL and LBC algorithms.

Based on the proposed distributed manner, each local LSTM model only needs to communicate with its neighbours. Thus, the STLF model can be trained locally and the multi-agent framework can reduce the computational and communication cost, which can be further extended to large networks \cite{dong2019demand}. Moreover, the distributed framework is more robust to single-point failures as long as the communication network remains connected \cite{dong2021short}.

\section{Case Study}\label{sec.simulation}
\subsection{Experiment and Model Setup}
\begin{figure}[thpb]
	\centering
	\includegraphics[scale=0.4]{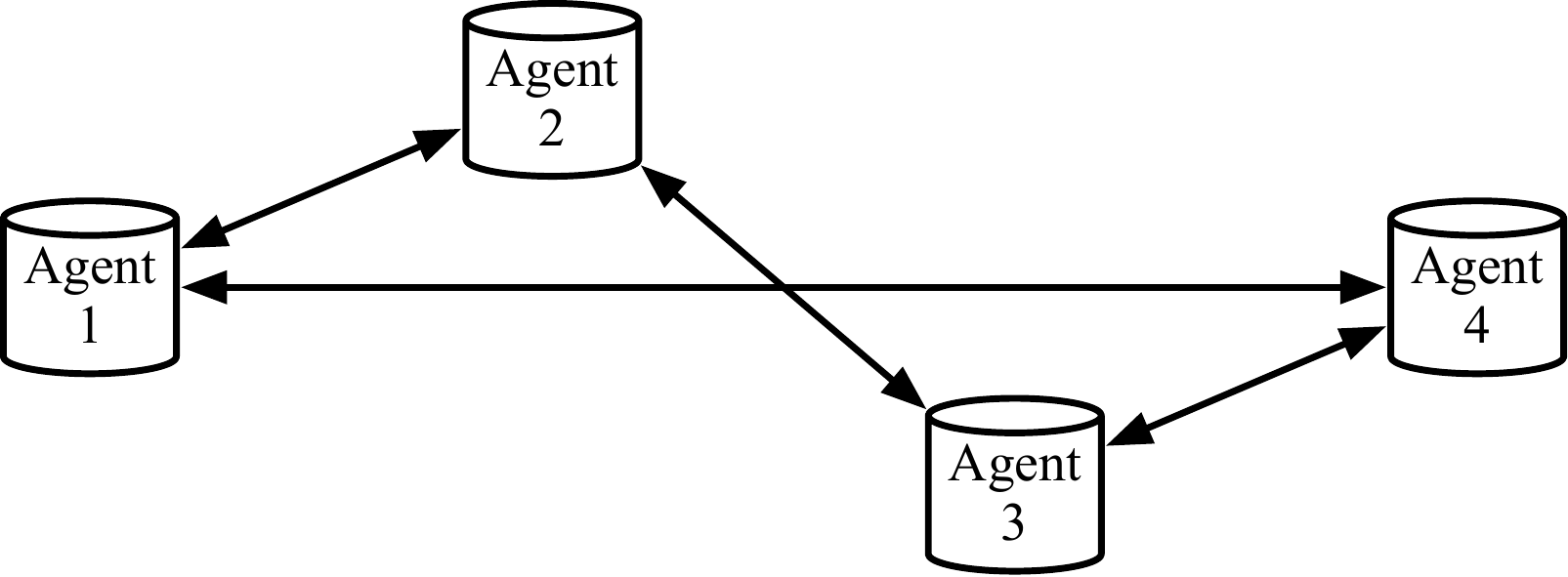}\vspace{-6pt}
	\caption{Communication Topology. \vspace{-6pt}}
	\label{communication topology}
\end{figure}

The case study is based on the historical load data, daily average temperature data and holiday type data for the period 2016-2019 which are provided by the GEFCom 2017 competition \cite{hong2016probabilistic} and ISO New England \cite{sahay2014day}. The system topology under consideration is the connection of four agents (data centres) as shown in Fig.~\ref{communication topology}, each of them has a quarter of dataset.
In addition, the iteration times of consensus between each information connection is set as 20, which makes the algorithm converge faster.

For STLF, the main influencing factors are historical load, temperature, date type and other variables. These three types of data are selected as input variables for the forecasting model. The load has a characteristic of a 24-hour periodic change, so that the power load value at the same time before the forecasting day and two days before can be used as the input variable of the model. In addition, considering the effects of temperature on the electrical load, daily mean temperature after preprocessing can be used as input variables for the LSTM. The input and output information designed in this paper are summarised in Table \ref{i_o information}.

\begin{table}[htbp]
\footnotesize
	\centering
	\caption{Input and output information.	\vspace{-6pt}}
	\label{i_o information}
	\begin{tabular}{cc}\hline
		input & input variables                                            \\\hline
		1-7                        & the electricity consumption of last week \\
		8-14                        & the average temperature of last week        \\
		15-21                        & the day type of last week            \\
		22                        & the electricity consumption at 2 days ago        \\
		23                        & the average temperature at 2 days ago       \\
		24                        & the day type at 2 days ago                   \\
		25                        & the actual load at 24 hours before   \\
		26                        & the average temperature at 24 hours before       \\
		27                        & the day type at 24 hours before              \\
		28                       & the temperature of the forecast day                \\
		29                       & the day type of the forecast day                   \\\hline
		output                   & the predicted load value at time $t$ of the forecast day         \\  \hline   
	\end{tabular}
\end{table}

Further, all raw data are pre-normalised to improve the accuracy of the forecasting model:
\small\begin{equation}
    x_i(k) = \dfrac{x_i(k) - x_i^{min}}{x_i^{max}-x_i^{min}}
\end{equation}\normalsize
where $x_i^{min}$ and $x_i^{max}$ are the minimum and maximum value of the $i$-th original input data. 
\subsection{Evaluation Criteria and Error Analysis}
The mean square error (MSE) and mean absolute percentage error (MAPE) are adopted as the final evaluating indicators, which are shown as follow:
\small\begin{equation}
	\varepsilon_{MAPE}=\dfrac{1}{N}\sum_{i=1}^{N}\left| \dfrac{y_{i}-\hat{y_{i}}}{y_{i}}\right| \times 100\%,
\end{equation}\normalsize
\small\begin{equation}
	\varepsilon_{MSE}=\dfrac{\sum_{i=1}^{N}(y_{i}-\hat{y_{i}})^{2}}{\sum_{i=1}^{N}(y_{i})^{2}},
\end{equation}\normalsize
where $y_{i}$ is the actual load at period $i$; $N$ is the total length of forecasting periods; and $\hat{y_{i}}$ is the forecast load at time $i$.

\subsection{Results and Analysis}

\subsubsection{Case 1}
\begin{figure}[htpb]
	\centering
	\includegraphics[width=\hsize]{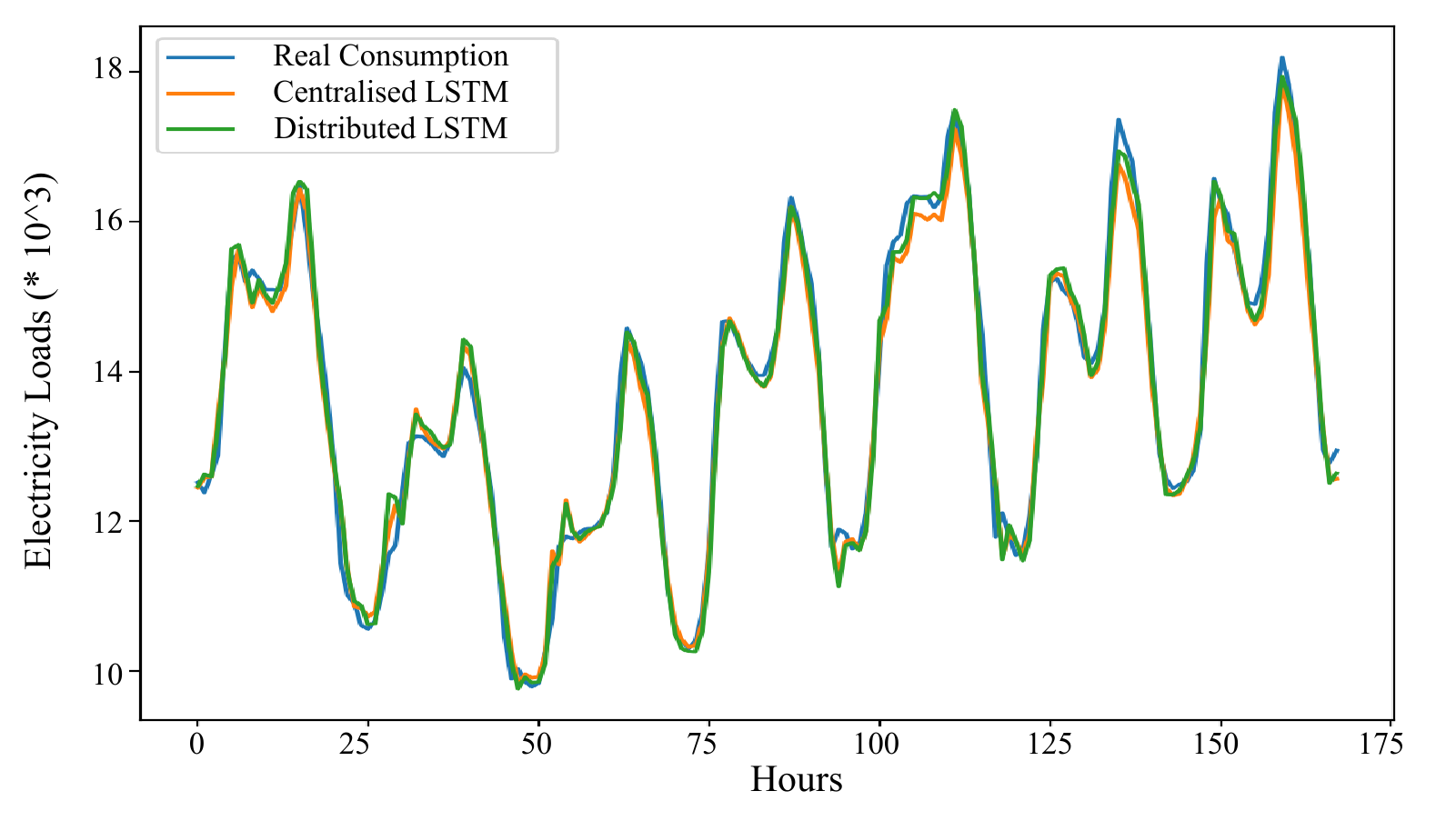}\vspace{-12pt}
	\caption{Short-term Load Forecasting Results.\vspace{-6pt}}
	\label{Prediction}
\end{figure}

This case verified the validity of the proposed DLSTM model. We use whole year customer electricity consumption data as the training dataset, and the test data is a dataset of a random week in next year. The results are shown in Fig. \ref{Prediction}. 

% \small 
% \begin{figure*}[htbp]
% \footnotesize
% \begin{floatrow}[3]
% \ffigbox{%
%   \includegraphics[width=\hsize]{Distributed STLF/figures/Prediction.png}
% }{%\centering
%   \caption{\centering \small Short-term Load\\ Forecasting Results.\vspace{-6pt}}
% 	\label{Prediction}
% }
% \ffigbox{%
% \centering
%   \includegraphics[width=\hsize]{Distributed STLF/figures/Prediction.png}
% }{%
%   \caption{\centering \small Distributed Training\\ Performance.\vspace{-6pt}}
% 	\label{TrainingPerformance}
% }
% \capbtabbox{%
%   \begin{tabular}{cccc}
% \hline
%       & MAPE       & MAE       & MSE               \\ \hline
% LSTM  & 0.0150 & 201.746 & 70021.160 \\
% DLSTM & 0.0138 & 183.352 & 62095.324 \\
% SVR   & 0.0152 & 202.246 & 73495.060 \\
% ANN   & 0.0280  & 324.650 & 131273.362    \\
% DBN   & 0.0206 & 144.326 & 44299.052 \\ 
% DANN   & 0.0275  & 284.240 & 123363.703    \\\hline
% \end{tabular}
% }{%
%   \caption{\centering \small Comparison of STLF\\ Models in Case 3.\vspace{-6pt}} \label{case3}%
% }
% \end{floatrow}
% \end{figure*}\normalsize

The green line is the forecasting results of the proposed DLSTM method, and it is evident that the proposed DLSTM has less prediction errors at most of the data points than centralised LSTM method. 
\subsubsection{Case 2}

\begin{figure}[htpb]
	\centering
	\includegraphics[width=\hsize]{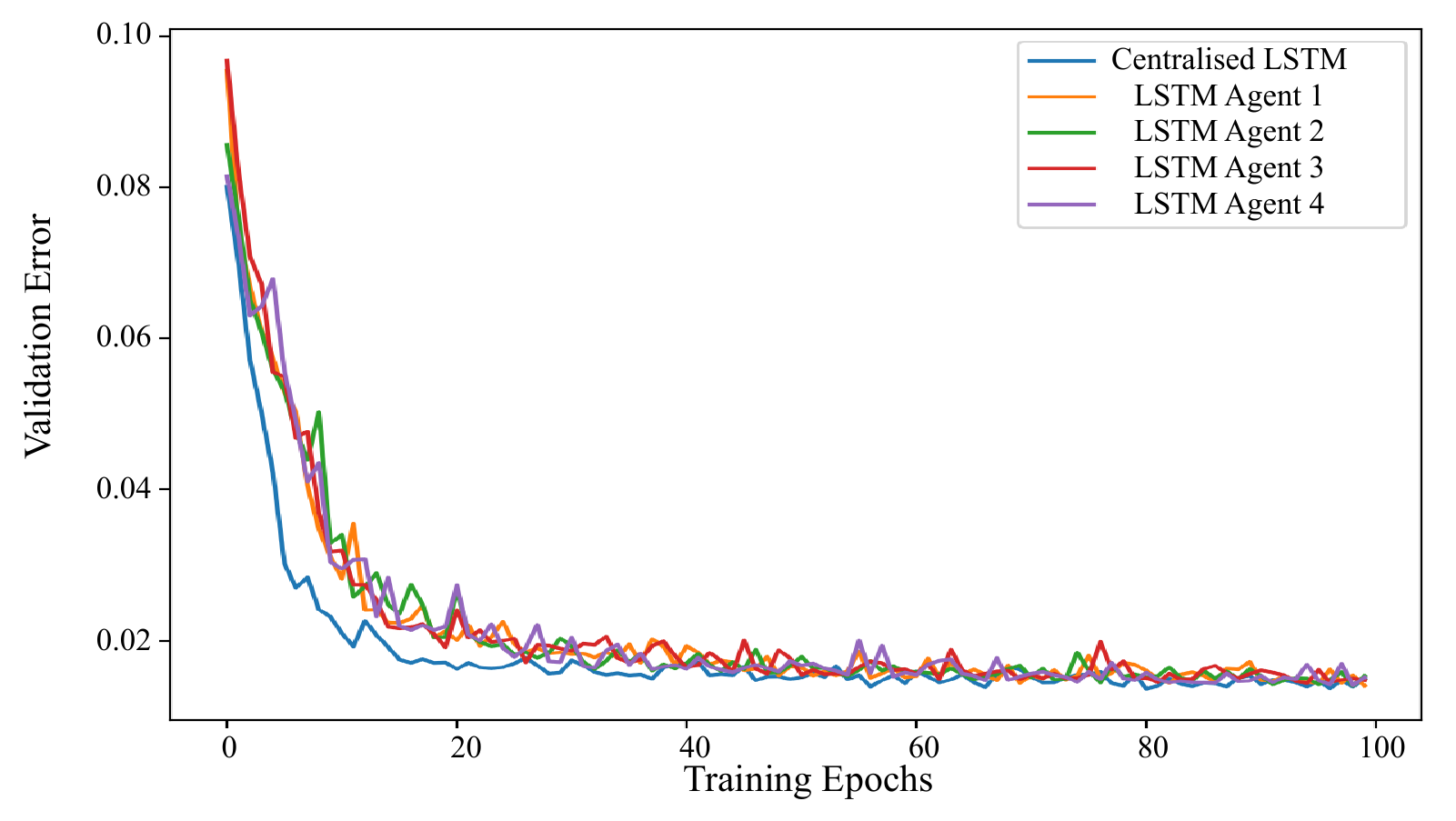}\vspace{-12pt}
	\caption{Distributed Training Performance. \vspace{-6pt}}
	\label{TrainingPerformance}
\end{figure}

In this case, we investigate the training performance of the proposed DLSTM model. Fig.~\ref{TrainingPerformance} depicts the validation errors during the training process among 4 distributed local DLSTM agents and centralised LSTM agent. Although each DLSTM agent only has a quarter of the dataset, it shows comparable training performance as the centralised LSTM method. 

In Fig.~\ref{TrainingPerformance}, we may clearly see that the converge speed of local computing LSTM models are slower than centralised LSTM model, but they can achieve similar accuracy after sufficient training epochs. Apart from the accuracy, the training speed of the proposed DLSTM is only 412 seconds, which is much faster than centralised LSTM (583 second). This is because that the dataset size of local training agent is only a quarter of centralised LSTM agent. 
Therefore, the proposed DLSTM algorithm can reduce the training time with an increasing number of computing agents, and the aim of fast training can be achieved by adding more computing agents if the load dataset continues to increase in the future.

\subsubsection{Case 3}
In this case, we summarise different STLF models to illustrate the advantage of the proposed model, and four typical state-of-the-art algorithms are chosen for comparison: \gls{SVR}, \gls{ANN}, \gls{DBN} and \gls{DANN}. The compared models are simulated based on the same dataset in this paper. All the models are trained with the same training dataset, validation dataset and test dataset. The results of different models are summarised in Table \ref{case3}.

% \begin{table*}[htbp]
% 	\footnotesize
% 	\centering
% 	\caption{\textcolor{black}{Comparison of Different STLF Models in Case 3.}}
% 	\begin{tabular}{cccccc}
% 		\hline
% 		     &   DLSTM       & DBN      & SVR     & LSTM   & DANN\\ \hline
% 		MAPE (\%) & $3.64  $    &$ 2.00$     &$ 1.98  $    & $2.44  $   & $3.79$\\
% 		MAE (W) & $460.53 $    & $202.31  $   & $199.61  $   & $256.33 $  & $478$\\
% 		MSE (W$^2$) & $3.36e^5$ & $7.27e^4$ & $7.19e^4$ & $1.13e^5$  & $3.63e^5$\\ \hline
% 	\end{tabular}
% 	\label{case3}
% \end{table*}
% \usepackage{graphicx}
\begin{table}[htbp]
\footnotesize
\caption{\textcolor{black}{Comparison of STLF Models in Case 3.}\vspace{-6pt}}
\centering
% \resizebox{\hsize}{!}{%
\begin{tabular}{cccc}
\hline
      & MAPE       & MAE       & MSE               \\ \hline
LSTM  & 0.0150 & 201.746 & 70021.160 \\
DLSTM & 0.0138 & 183.352 & 62095.324 \\
SVR   & 0.0152 & 202.246 & 73495.060 \\
ANN   & 0.0280  & 324.650 & 131273.362    \\
DBN   & 0.0206 & 144.326 & 44299.052 \\ 
DANN   & 0.0275  & 284.240 & 123363.703    \\\hline
\end{tabular}%

% }
\label{case3}
\end{table}

From the results provided in Table. \ref{case3}, the DLSTM algorithm meets the requirements of STLF and has better accuracy than other models. Comparing with the distinct machine learning algorithms, the LSTM and DLSTM reveals better performance and less prediction errors due to the time-sequence characteristic of STLF problems. 
For distributed algorithms, we can see that the DANN and DLSTM show the similar accuracy as ANN and LSTM, respectively.
It can be seen from the Table. \ref{case3} and Fig. \ref{TrainingPerformance}, the proposed DLSTM method demonstrates similar and even better accuracy than traditional single centralise LSTM algorithm after around 50 training epochs. %Therefore, the proposed DLSTM algorithm can benefit the STLF especially under the big data scenario.}

\section{Conclusion and Future Work}
\label{sec_con}
This paper ascertains the effectiveness of using the distributed long short-term memory models in short-term load forecasting. The multiple and time-variable variations can be predicted by the proposed approach. The decentralised protocol makes it possible to separate data sets and pre-train models at an individual data centre, which simplifies the forecasting problem and leads to more privacy results. In the application examples, the proposed distributed long short-term memory model provided accurate forecast results and faster training speed than other machine learning models. The distributed model is simpler to train and tune than the centralised models, does not over-fit and reduces variance due to the consensus weight of many subsets. In future, we plan to apply trustworthy AI techniques, e.g. \cite{huang_coverage_guided_2021,zhao_safety_2020,zhao_baylime_2021}, to assure the reliability and security of our method, as well as its explainability.

% \bibliographystyle{IEEEtran} 
% \bibliography{reference}

\end{document}